\documentclass[letterpaper, 10 pt, conference]{ieeeconf}      

\IEEEoverridecommandlockouts                              

\usepackage{booktabs}
\usepackage{cite}
\usepackage{multicol}
\usepackage{booktabs}
\usepackage{tabularx}
\usepackage{graphicx}
\usepackage{cuted}
\usepackage{capt-of}
\usepackage{amsmath,amssymb}
\DeclareMathOperator*{\argmax}{arg\,max}
\usepackage{tcolorbox}
\usepackage{algorithm}
\usepackage{algpseudocode}

\usepackage[bookmarks=true]{hyperref}
\usepackage{subfiles} 

\usepackage{xcolor}
\usepackage{xstring}
\let\oldtextcolor\textcolor
\renewcommand{\textcolor}[2]{%
  \IfEqCase{#1}{%
    {red}{}
  }[\oldtextcolor{#1}{#2}]
}

\title{\LARGE \bf
SyncSBC: Decentralized Swarm Behavior Prediction for Synchronized Autonomous Control
}

\author{
Varun Raveendra$^{1}$, Connor Mattson$^{1}$, Daniel S. Brown$^{1}$ %
\thanks{$^{1}$Kahlert School of Computing, University of Utah, Salt Lake City, USA }%
}

\begin{document}

\maketitle
\thispagestyle{empty}
\pagestyle{empty}

\begin{strip}\centering
\vspace{-3em}
\includegraphics[width=\textwidth]{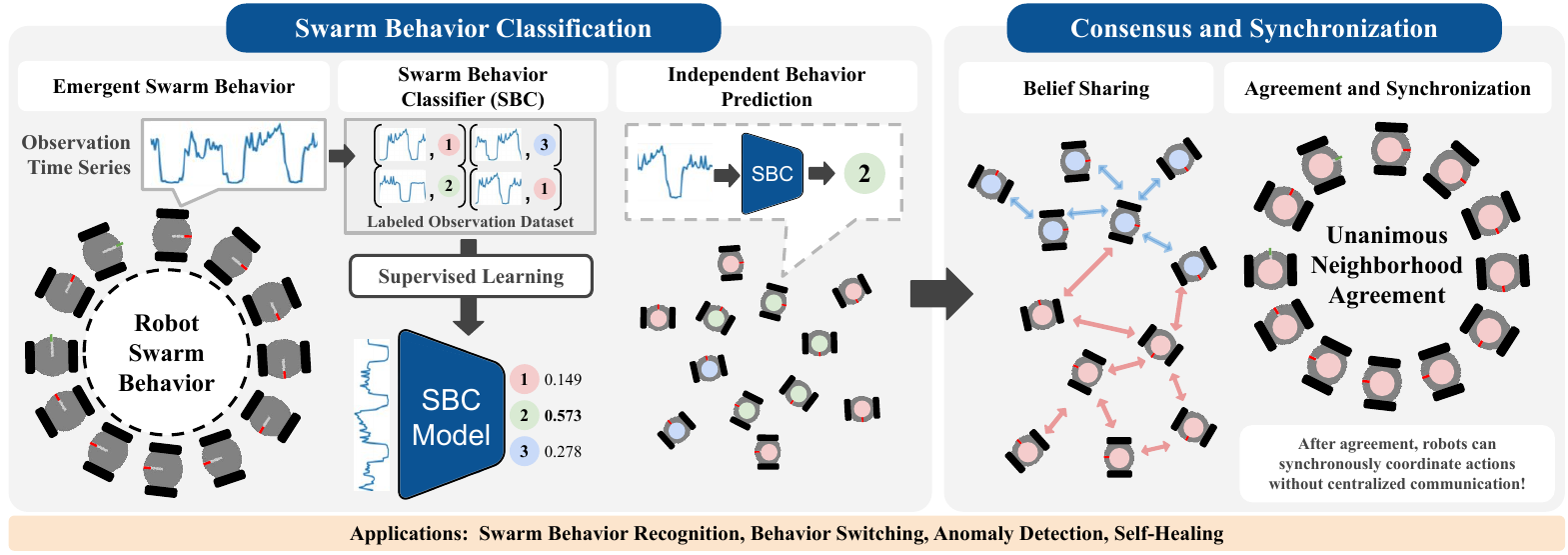} 
\captionof{figure}{\textbf{Synchronous Swarm Behavior Classification (SyncSBC)} is a dual-stage fully-decentralized framework for transforming agent-level behavioral predictions into synchronized action execution using limited-range communication and consensus. (\textbf{Left}) A trained deep neural network predicts the behavioral properties of the swarm using only egocentric local observations. (\textbf{Right}) A consensus algorithm measures neighborhood agreement and synchronizes executed actions. Example applications for SyncSBC include behavior switching and anomaly detection, which we highlight in our experiments.  
\label{fig:teaser}}
\end{strip}

\begin{abstract}
Robot swarms utilize many independent limited-sensing agents to produce complex emergent behaviors without requiring centralized control. 
However, little research explores how agents can infer swarm-level behavior from purely local perception, a capability critical for detecting faults and behavior changes.
In this paper, we introduce Synchronized Swarm Behavior Classification (SyncSBC), which combines improvements in machine learning and distributed consensus to classify collective swarm behavior and synchronize swarm decision-making in an entirely decentralized manner. 
We show that SyncSBC achieves high classification accuracy and low synchronization delay, making it suitable for real-world deployment.
Finally, we use SyncSBC to demonstrate two promising swarm applications on real robots
where we show that swarms utilizing SyncSBC can accurately identify anomalies in robot behavior and autonomously coordinate collective changes in swarm behavior. Videos, code and supplemental experiments are available at \url{https://sites.google.com/view/sync-sbc/home}.
\end{abstract}

\section{Introduction}

Multi-agent robotic systems empower scientists and engineers to solve problems using teams of agents. While many real-world systems incorporate highly capable robots, advanced communication networks, and perception pipelines to inform agent decisions, they may be infeasible for development and scaled deployment due to hardware, space, and budget constraints~\cite{KIM2026100841}. By contrast, swarm robotics enables tasks such as agriculture monitoring~\cite{blender_managing_2016}, underwater surveillance~\cite{duarte_evolution_2016}, search and rescue~\cite{arnold_search_2018}, and object transportation~\cite{wilson_multi-robot_2018} to be solved using robots with limited communication and sensing capabilities, often with agents operating under a partially observable view of the task and other agents. However, this decentralization creates a challenge for agents when they must make locally informed decisions that affect global behavior while having only partial visibility of the swarm's collective state. Furthermore, due to limited observability, agents typically are unaware of faulty or unstable agents in the swarm, which can further degrade collective performance~\cite{faulty}. 

We seek to enable individual swarm agents to coordinate and estimate global, emergent collective behaviors from local sensing and communication. This has the potential to improve collective decision-making, autonomous swarm control, and structured orchestration within the swarm. 
However, introducing communication in swarms has its own challenges. Message exchange grows with the system size, increasing overhead and potentially leading to network congestion and stability issues~\cite{message,message2}. Thus, methods for global behavior detection must operate under strict communication rates. This motivates the following question: \textit{How can we provide individual agents with the means to accurately detect and adapt to swarm-level behavior using local perception and limited decentralized communication?}

Prior work has explored learning-based approaches to algorithmically uncover control methods and parameters that, when deployed on swarm robots, efficiently produce a desired swarm-level behavior~\cite{ozdemir2017shepherding}. But do not address how individual agents can classify global behavior from local perception. 
Other swarm methods focus on global behavior agreement through local interactions, using majority rules and voter models~\cite{val}, Kalman Consensus Filters~\cite{olfatisaber2007distributed}, and Event-triggered consensus~\cite{NOWZARI20191}. These methods ensure distributed state alignment under known/predefined decision variables. 
Additionally, prior work has studied how the behavior of a swarm changes in response to environmental conditions~\cite{yang2022autonomous} or robot failures~\cite{timmis2016immune}, emphasize robustness, yet do not address decentralized estimation or stabilization of behaviors.  
By contrast, our paper focuses on learned global behavior estimation through local perception and detecting when these estimates stabilize across agents. We combine this with a lightweight synchronization mechanism for discrete commitments while minimizing communication.

In this paper, we introduce Swarm Behavior Classifier (SBC), a novel method that leverages machine learning to train a shared classifier model using local observations from individual agents to predict the global collective behavior of the swarm. Combining this with a hybrid consensus mechanism that incorporates synchronized consensus update to a behavior across all agents through communication, we present our method Synchronized SBC (SyncSBC). SyncSBC, as shown in Figure~\ref{fig:teaser}, allows individual agents in a swarm to map their local observations into temporally aligned conclusions of the collective behavior observed by the swarm. Utilizing this method, we demonstrate two use cases that have promising applications in real-world tasks: Behavior Switching and Anomaly Detection.

We demonstrate the efficacy of our approach through simulation and on real robots by systematically evaluating machine-learning models and consensus methods at each stage. We present inference performance of different ML architectures and demonstrate how our hybrid consensus method can minimize inter-robot synchronization delay. Furthermore, we demonstrate SyncSBC's ability to detect behavioral convergence and leverage it as a signal to trigger subsequent control actions. We also demonstrate SyncSBC's inherent ability to detect faulty or anomalous robots in the swarm with high confidence. 

Our work can be summarized with the following contributions: \textbf{(1)} We present a novel decentralized framework for online classification of global swarm behavior using an individual agent's partially observable sensor signals, namely Swarm Behavior Classification (SBC).
\textbf{(2)} We propose a novel distributed behavior consensus and convergence synchronization pipeline by combining event-triggered communication~\cite{NOWZARI20191}, and coordination mechanisms~\cite{booleang,mahy2004preserving}, leveraging decentralized communication to synchronize agent consensus across the swarm. 
\textbf{(3)} We show that SyncSBC is a robust method for deployed robot applications such as  anomaly detection, and closed-loop switching of behaviors in real time on real robots. 
\section{Related Work}

\textit{Swarm Behavior Detection:} 
Using external sensors and cameras to track the behavioral characteristics of swarms has been a well-studied method that enables researchers to evaluate behavior online in real time or post hoc offline ~\cite{wu2022swarm}. Prior work has studied using compressive subspace learning~\cite{berger2016classifying}, conformal prediction~\cite{xi2023conformal} and Bayesian classification~\cite{brown2014limited} to enable \textit{centralized} swarm behavior prediction and classification.
In contrast to prior work \textcolor{black}{that forms beliefs over predefined global hypotheses~\cite{Zheng2023EntropyConsensus} or predefined alternatives~\cite{article33}, }we are, to the best of our knowledge, the first to explore and propose an approach for \textit{decentralized} swarm agents to classify emergent \textcolor{red}{swarm behavior} \textcolor{black}{collective dynamics} using only locally observable information from their sensor data.  

\textit{Consensus Methods:} Consensus decision-making can be classified into two categories, one where agents need to choose from infinite options, and the other where robots have finite options. The latter category is commonly referred to as the Best-of-M problem~\cite{bom}. Best-of-M problems can be classified as symmetric~\cite{inbook}, where all options are equal, or asymmetric~\cite{assymetry} if unequal. Best-of-M has been studied for M=2, where swarms estimate the dominating option~\cite{article33}. For M$>$2, probability sharing approaches are used~\cite{probab}. Valentini et al.~\cite{val} proposed voter-based methods in collective consensus for perception. While these approaches achieve agreement over predefined options, they primarily focus on convergence and do not address temporal synchronization of decisions across agents. In this paper, our goal is to employ distributed decision-making strategies that ensure swarm alignment and extend to concepts of synchronization and its importance in swarms with dynamic decision states. Our setup solves a symmetric, homogeneous, time-varying distributed synchronization problem, unlike classical consensus mechanisms~\cite{olfati2}.

\begin{figure*}[h]
    \centering
    \includegraphics[width=0.9\linewidth]{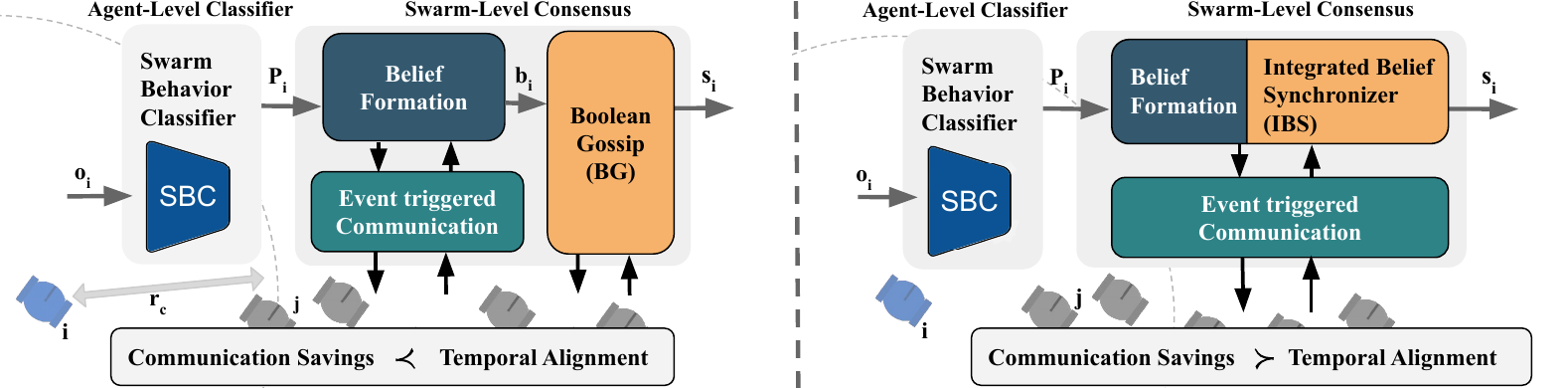}
    \caption{\textbf{Synchronization and Communication in SyncSBC.} We study two consensus methods for use in our framework based on system design preferences: (\textbf{Left}) when minimal synchronization delay is desired, we propose using event-triggered communication and boolean gossip and (\textbf{Right}) when reducing communication messages is preferred, then a combination of event-triggered communication and integrated belief synchronizer is optimal.}
    \label{fig:pipeline}
\end{figure*}


\section{Problem Statement}\label{ps}

\textcolor{red}{Robot swarms are typically capable of exhibiting different emergent collective behaviors, which we call behavior classes. 
Given a swarm of $N$ robot agents and a set of behavior classes, $C = \{ c_1,..., c_n\}$, we first aim to train a model that can classify the collective behavior of the swarm using only a finite history of the last $T$ observations, $O_{i, t} = (o_{i, (t-T)}, \cdots, o_{i, (t-1)}, o_{i,t})$ for any given robot $i\in N$. 
We learn a parameterized model $\phi: O_{i, t} \to C$ mapping local observations to class probabilities $p_i(t)$.}

\textcolor{black}{Let $\mathcal{V}=\{1,\ldots,N\}$ represent the set of robot agents and let $\mathcal{C}=\{c_1,\ldots,c_M\}$ denote the set of behavior classes. Robot $i\in\mathcal{V}$ makes a local observation $\mathbf{o}_i(t)\in \mathbb{R}^{d_o}$ and a history of local observations of length $T$, $\mathbf{O}_{i,t} =\left(\mathbf{o}_i(t-T+1),\ldots,\mathbf{o}_i(t)\right)\in\mathbb{R}^{T\times d_o}$. We learn a shared model that can classify the collective behavior of the swarm $\phi:\mathbb{R}^{T\times d_o}\rightarrow \Delta^{M-1}$, where $\Delta^{M-1}=\{\mathbf{p}\in\mathbb{R}_{\geq0}^{M}
\mid \mathbf{1}^{\top}\mathbf{p}=1\}$
ensures that the $M$ class
probabilities are nonnegative and sum to one, yielding
$\mathbf{p}_i(t)=\phi(\mathbf{O}_{i,t})$.}

\textcolor{red}{We assume access to training data in the form of sensor readings for all agents in the swarm when executing stable behavior modes, each associated with a behavior class. Stable behavior modes refer to swarm episodes that exhibit consistent observable patterns (e.g., Fig.~\ref{fig:robot}).
The set of all $N$ agent's local observation histories in an episode is
denoted as $E = (O_{1, t}, \cdots, O_{N, t})$ that consists of a sequence of fixed-length observation histories. 
This results in a dataset of stable behavior episodes, each with the ground-truth behavior label, $y$, corresponding to the true resulting collective behavior of the swarm, $D = \{(E_0, y_0), \cdots, (E_m, y_m) \}$.}

\textcolor{black}{Training data are collected from $R$ stable-behavior episodes that exhibit consistent observable patterns (e.g., Fig.~\ref{fig:robot}). Episode $e\in\{1,\ldots,R\}$ contains, for each robot $i\in\mathcal{V}$, a local observation sequence
$\{\mathbf{o}^{(e)}_i(t)\}_{t=1}^{L_e}$ and an episode-level
class label $\ell_e\in\{1,\ldots,M\}$. Every local observation window from every robot inherits an episode label that corresponds to the resulting collective behavior of the swarm, yielding the agent-level training dataset $\mathcal{D}=\{(\mathbf{O}^{(e)}_{i,t},\ell_e)\mid
e\in\{1,\ldots,R\}, i\in\mathcal{V}, t\in\mathcal{T}_e\}$ where $\mathcal{T}_e$ contains window endpoints within the stable portion or the portion with consistent patterns (Fig.~\ref{fig:robot}) of episode $e$. }

\textcolor{red}{We study the case where each robot $i \in \{1, \dots, N\}$ communicates with neighboring robots within a bounded radius $r_c$. Let $x_i(t) \in \mathbb{R}^d$ denote the position of robot $i$ at time $t$. The neighbor set is defined by \[
\mathcal{N}_i(t) =\{j \in \{1, \dots, N\}
\;\\;
\|x_i(t) - x_j(t)\| \le r_c,\; j \neq i
\}
\]
This is a dynamic communication graph $\mathcal{G}(t)$ where the topology changes as the robots move in the environment~\cite{Zheng2023EntropyConsensus}.   }

\textcolor{black}{We study the case where each robot $i$, at position $\mathbf{x}_i(t)\in\mathbb{R}^d$, communicates with neighbors within radius $r_c$, defining
$
\mathcal{N}_i(t)=\{j\in\mathcal{V}\setminus\{i\}\mid
\|\mathbf{x}_i(t)-\mathbf{x}_j(t)\|_2\le r_c\}
$
and the time-varying graph $\mathcal{G}(t)$~\cite{Zheng2023EntropyConsensus}. This graph $\mathcal{G}(t)$ remains connected, such that information can propagate between
all robots through direct or multi-hop communication.}

\textcolor{red}{We aim to utilize a decentralized consensus mechanism over the time-varying ground-truth label $y$ to enable the swarm to efficiently align on the observed collective behavior through local communication. Internal belief state $b_i(t)$ represents the local agreement state for agent $i$ at time $t$ achieved during consensus update. Consensus at time $t$ is achieved if $b_i(t) = y$ for all $i$. 
An agent-level true positive $b_i(t)=y$ is when $t \ge t_{\mathrm{GT}}$, where $t_{\mathrm{GT}}$ denotes the time at which the swarm exhibits a stable behavior mode.}

\textcolor{red}{For each agent $i$ let $t^*$ be the time at which agent sets its synchronized belief $s_i(t_i^*)=y$. 
True swarm consensus is achieved when $s_i{(t)=y}$ for all agents $i$. Temporal alignment is quantified by the \textit{synchronization delay}
$\Delta = \max_i t_i^* - \min_i t_i^*$, where $t_i^*$ refers to the time at which agent $i$ sets its synchronized belief.}

\textcolor{black}{During deployment, the collective behavior may change over time. Let
$\ell(t)\in\{1,\ldots,M\}$ denote the ground-truth class index and
$\mathbf{y}(t)\in\{0,1\}^{M}$ its one-hot encoding. Each robot maintains
an internal belief $\mathbf{b}_i(t)\in\{0,1\}^{M}$ and a synchronized
belief $\mathbf{s}_i(t)\in\{0,1\}^{M}$.}

\textcolor{black}{For a stable behavior interval beginning at $t_{\mathrm{GT}}$, let
$\mathbf{y}^{*}$ denote the one-hot ground-truth behavior. Robot
$i$ forms the correct internal belief at
$t_i^{b}=\min\{t\mid\mathbf{b}_i(t)=\mathbf{y}^{*}\}$. Since robots rely
on local observations, $t_i^{b}$ may differ across robots; an internal
belief formed before $t_{\mathrm{GT}}$ is considered a false positive.
Let
$t_i^{*}=\min\{t\geq t_{\mathrm{GT}}\mid
\mathbf{s}_i(t)=\mathbf{y}^{*}\}$
denote the time at which robot $i$ forms the correct synchronized belief.
Synchronized swarm consensus is achieved when
$\mathbf{s}_i(t)=\mathbf{y}^{*}$ for all $i\in\mathcal{V}$.
Temporal alignment is quantified by \textit{synchronization delay}
$\Delta=\max_{i\in\mathcal{V}}t_i^{*}
-\min_{i\in\mathcal{V}}t_i^{*}$, where $t_i^*$ is the time at which agent $i$ sets its synchronized belief.}

Our goal here is to first learn a function $\phi$ through deep function approximation, that maps local observations from individual agents $\mathbf{O}_{i,t}$ to predictions $\mathbf{p}_i(t)$, and then utilizing the graph $\mathcal{G}(t)$ to efficiently communicate and employ a synchronizing consensus operator $\mathcal{H}_\theta$ by optimizing its parameters $\theta$. Where $\mathcal{H}_\theta$ maps $(\mathbf{p}_i(t),\mathbf{b}_j(t),\mathbf{b}_i(t))$ to $\mathbf{s}_i(t)$ for agent $i$ and neighbor $j$ in set $\mathcal{N}_i(t)$ while minimizing $\Delta$.

\section{Agent-Level Classification}
To enable agent-level predictions of collective swarm behavior, we introduce the Swarm Behavior Classifier (SBC) Model shown in Figure~\ref{fig:teaser}. In this classifier, each agent utilizes its locally available observations to infer the global swarm behavior. In contrast to global vision-based tracking methods, which require access to high-resolution camera trackers or other methods that require full knowledge of the swarm (robot state), SBC operates without a central system or full knowledge of the swarm, making it more scalable and practical in real-world, unstructured environments. 

We propose training a global behavior classifier solely from local observations of individual agents. By training on time-varying local sensor streams, we seek to capture dynamical patterns that reflect the swarm's collective behavior and enable local estimates of the global behavior classes. Swarm robots can be equipped with range sensors, time-of-flight sensors, LiDAR, or cameras that provide local measurements of objects or environmental features in their vicinity. These time-series signals are well-suited for processing with neural network architectures such as convolutional models, which we emphasize in our experiments, but this approach can generalize to alternate architectures and input modalities.
\textcolor{red}{The SBC model is then trained over the space of predefined candidate behaviors, with appropriate ground truth labels using supervised learning. This model is shared across all robots and enables each robot to map its raw sensor history to a probability distribution over behaviors, denoted as $p_i$.}\textcolor{black}{The SBC model is then trained over the space of predefined candidate behaviors, with appropriate ground truth labels using supervised learning. Each robot applies the model $\phi$ to its local
observation history $\mathbf{O}_{i,t}$, yielding class probabilities
$\mathbf{p}_i(t)=\phi(\mathbf{O}_{i,t})$.}
\textcolor{red}{To mitigate noise and fluctuations in predicted behaviors, we apply the Exponential Weighted Moving Average (EWMA) as a temporal smoothing operator. The EWMA update rule is given by}\textcolor{black}{Predictions are smoothed using Exponential Weighted Moving Average (EWMA)}: $P_i(t) = P_i(t-1)+ \alpha (p_i(t) - P_i(t-1))$\textcolor{red}{. Here, $P_i(t)$ represents the smoothed prediction probability vector at step $t$ for each behavior class, and $\alpha$ is the smoothing factor that controls the weight assigned to newer predictions from $p_i(t)$.}\textcolor{black}{, where $\alpha$ is the smoothing coefficient.} 

In the following section, we assume access to a pretrained SBC model and discuss how to use this model for collective behavior consensus. We describe our specific instantiation of the SBC model in Section~\ref{se}.


\section{Synchronized Belief formation}

Individual agents compute egocentric estimates of the collective behavior using SBC, but these estimates are often misaligned over time with the swarm's ground-truth behavior and are prone to false positives. To better understand ground-truth behavior, swarm agents can leverage consensus methods to assess prediction agreement with neighboring agents. 

In a decentralized swarm setting, individual agents lack access to the full global state and must rely on local sensing and limited communication mechanisms.  To address this limitation, we propose using a distributed consensus model that enables robots to broadcast their local predictions within a communication radius $r_c$. We decompose our synchronized belief formation method into two stages: Internal Belief Formation, for local belief updates, and Synchronization, which aligns belief formation through temporal alignment.

\subsection{Stage-1: Internal Belief Formation}\label{s1}
After each robot in the swarm produces its initial estimate $P_i(t)$ of the collective behavior, the robots can now enter the communication stage where they map local predictions $P_i$ to an internal belief $b_i$ that shows agreement to a global behavior based on alignment with its neighbors. The internal belief $b_i$ is a per-class binary vector for each agent $i$.
Formally, internal belief formation solves the best-of-M problem where agents agree on the same behavior predicted class while communicating with their neighbors. 

Stage 1 requires us to apply consensus mechanisms to the agent's local prediction to realize internal belief estimates $b_i$. In this work, we consider a few candidate strategies to determine which strategy closely tracks the ground-truth collective behavior:

\textbf{Averaging:} Each robot performs nearest-neighbor average consensus over its internal belief, where it updates its belief as the mean of its own predictions and its neighbors~\cite{olfati2}. 

\textbf{Entropy-based Fusion:} Following~\cite{Zheng2023EntropyConsensus}, robots share their strongest choice from the softmax predictor values, along with an entropy of their distribution, representing a measure of certainty about the choice they shared. 

\textbf{Neighbor Variance Update:} A method inspired by the disagreement measure~\cite {olfati2}, where the selected class is the one with the lowest disagreement, 
$\left( P_{i,k}(t) - P_{j,k}(t) \right)^2$,
where $P_{i,k}$ denotes the predicted probability assigned by agent $i$ to behavior $k$, and $P_{j,k}$ \textcolor{black}{over all $j \in \mathcal{N}_i(t)$.}

\textbf{Sample and Hold Strategy:} An update mechanism inspired by event-triggered control~\cite{sah}, where agents $i$ update their internal estimates $I_i(t)$ when self-disagreement $\left(P_i(t) - I_i(t)\right)^2$ exceeds a predefined threshold. An agent's internal belief $b_i$ is set when a selected class is stable and has low disagreement.

In all cases, each robot $i$ propagates its strategy-specific messages $\boldsymbol{\mu}_i$ to its neighbors in set $\mathcal{N}_i$, which consists of robots within the communication radius $r_c$. 
These strategies operate only to determine an agent's internal belief update $b_i$ and are designed to be modular, allowing alternate consensus strategies to be integrated beyond those mentioned above based on application context.



\subsection{Stage-2: Synchronization}\label{s2}
Distributed consensus methods facilitate agreement among neighboring agents, but do not inherently focus on temporal synchronization across the swarm in the absence of a centralized controller, leading to misalignment and fragmented decision regions.
Synchronization~\cite{message} is essential for reliable swarm deployment, particularly for tasks that require coordinated transitions or closed-loop control. In this second stage, we aim to ensure that all neighbors are synchronized for a swarm-level consensus, minimizing the synchronization gap $\Delta$ defined in Section~\ref{ps}.  To achieve this, we introduce two synchronization methods: Boolean Gossip and Integrated Belief Synchronizer as shown in Figure~\ref{fig:pipeline}. Each method has its own trade-offs in convergence time and robustness.

\textit{Synchronization using Boolean Gossip (BG):}  To realize synchronized belief $s_i$ from internal belief states, we employ a Boolean gossip mechanism pictured in Figure~\ref{fig:pipeline}~(left). Boolean gossip propagates individual internal beliefs $b_i$ through neighbor interactions, similar to monotone epidemic-style information dissemination in randomized gossip algorithms~\cite{booleang}. Each robot maintains a transient synchronized belief variable $s_i$, which is iteratively updated via a logical AND operation over its internal binary belief state and a transient synchronized belief state received from its neighbors, $s_i=b_i \land s_j$. Robots additionally broadcast their states with a per-round sequence counter, enabling neighbors to align to a common epoch~\cite{lamport1978time}.

\textit{Synchronization using Integrated Belief Synchronizer (IBS):}  We introduce a novel message-piggybacked synchronization mechanism operating over $k$ behavior-specific fields. Each field $\mathcal{F}_k=(\mathbf{\tau}_k,e_k,\mathcal{M}_k)$ maintains a distributed state consisting of: 
1) Token owner $\mathcal{\tau}_k$, 2) Lamport-style epoch $e_k$~\cite{lamport1978time}, and 3) Membership set containing the robots whose internal belief $b_i$ supports that behavior $\mathcal{M}_k=\{i|b_i=k\}$.

At each update, robots merge received neighbor states by selecting the state with the highest epoch $e_k$, ensuring resolution of concurrent updates $\mathcal{F}_{i,k} \leftarrow \arg \max_{\mathcal{F}_{j,k}} e_k $ ~\cite{suzuki1985distributed, mahy2004preserving} and preventing split-decision formation. A robot may modify the behavior's membership $\mathcal{M}_k$ of its most confident behavior in $b_i$ only if the token $\tau_k$ is free or locally held, thereby enforcing controlled updates. Synchronized belief $s_i$ is set once all robots in the neighborhood agree and no active token holders remain, enabling a stable and decentralized synchronization layer we call the Integrated Belief Synchronizer (IBS) visualized in Figure~\ref{fig:pipeline}~(right).

While the underlying primitives in IBS and Boolean Gossip are inspired by techniques in distributed systems~\cite{birman1987reliable}, their integration into decentralized swarm-level consensus and synchronization under dynamic communication topologies constitutes the system-level contribution of SyncSBC.

\subsection{Communication Reduction}
We further incorporate Event-Triggered Communication (ETC)~\cite{NOWZARI20191} which replaces inter-robot periodic message transmissions with condition-triggered transmissions. This reduces redundant transmissions and alleviates network congestion in swarm deployments, promoting scalability. This triggering rule follows the event-triggered consensus framework~\cite{sah}. In ETC, agents maintain an internal state initialized as $I_i(0)= P_i(0)$, we measure internal disagreement $E_i(t)=\left\|P_i(t) - I_i(t)\right\|^2$ and local disagreement with neighbors $n_i(t)=\sum_{j \in \mathcal{N}_i(t)}\left\|I_i(t) - I_j(t)\right\|^2$. The internal state is updated $I_i(t)= P_i(t)$ and relevant messages are transmitted only when condition $ E_i(t)\;\ge\;\sigma \, n_i(t)$ is satisfied where $\sigma$ is a hyperparameter $>0$.

ETC enables controlled message propagation. Pairing it with the Boolean gossip algorithm results in lesser communication overhead; however, pairing ETC with IBS yields a substantially greater reduction. This is because Boolean Gossip relies on iterative message propagation for synchronization, and IBS does not require periodic propagation.

\begin{figure}[t]
    \centering
    \includegraphics[width=1.0\linewidth]{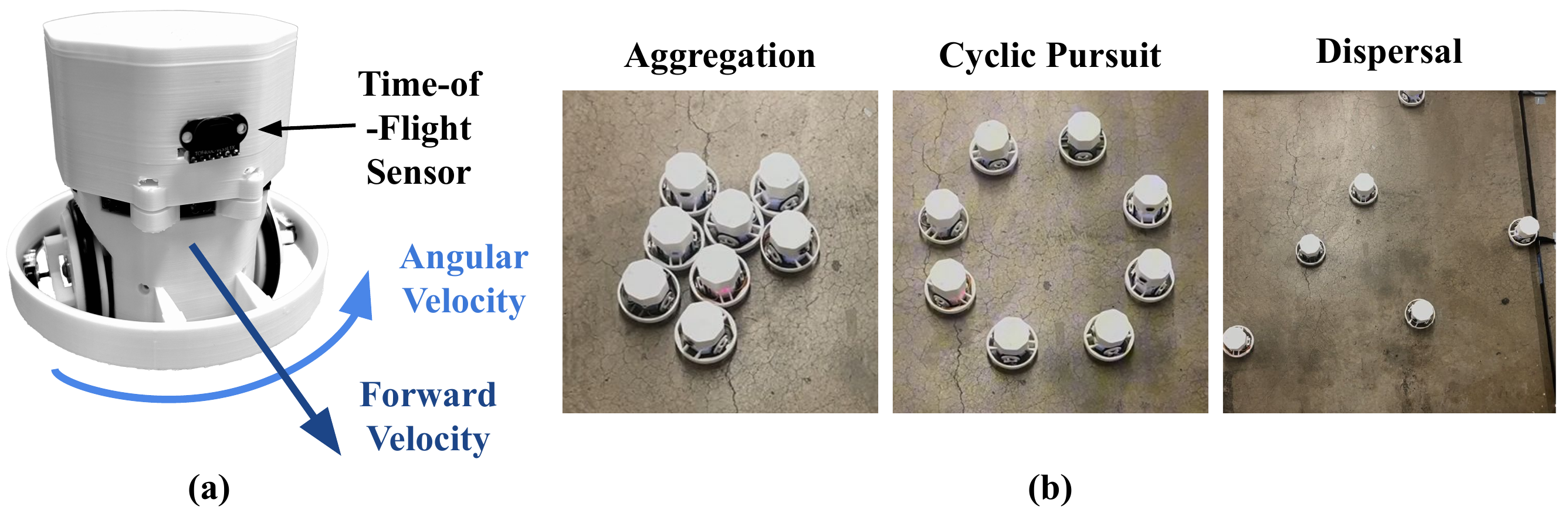}
    \caption{\textbf{HeRo+ Robots:}
     \textbf{(a)} A single HeRo+ robot uses unicycle commands to locomote and time-of-flight sensing to detect other robots. \textbf{(b)} 3 emergent behaviors deployed on 8 robots. From Left to Right: \textit{Aggregation}, where agents aggregate at the swarm's centroid, \textit{Cyclic Pursuit}, where agents travel about the centroid in a circular motion, and \textit{Dispersal}, where agents maximize the inter-agent distance to span (explore) the environment. }
    \label{fig:robot}
\end{figure}

\section{Simulated Experiments}\label{se}

We evaluate SyncSBC across multiple pipeline stages and present results from simulated experiments conducted on swarm robots. Our swarm robots have limited sensing and actuation capabilities, with line-of-sight sensors and differential-drive wheel motors controlled by linear and angular velocities. The swarm controllers we study are defined as linear and angular velocity commands that depend on whether a robot detects another robot in its line of sight. Specifically, controllers are parameterized as ($V_t, a_t, V_f, a_f$) where $V_t$ and $a_t$ correspond to linear and angular velocity when a robot is detected in its line-of-sight and $V_f$, $a_f$ when not detected.
In our work, we study the following swarm behaviors using controllers published in prior work~\cite{mattson2025discovery}, which exhibited distinct collective dynamics and were shown to work on low-cost, open-source robots: \textit{Cyclic Pursuit}, \textit{Aggregation}, and \textit{Dispersal}. Depictions and descriptions of these behaviors can be found in  Figure~\ref{fig:robot}b.




\begin{figure}
    \centering
    \includegraphics[width=\linewidth]{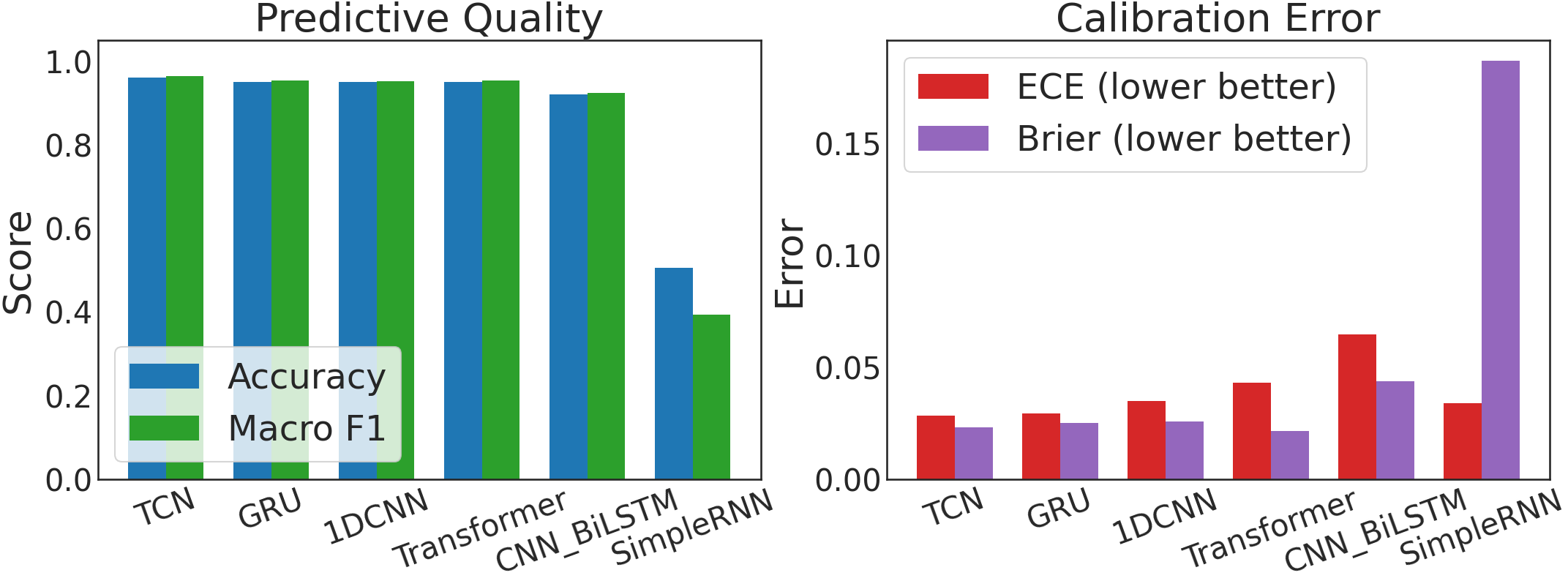}
    \caption{
    \textbf{Agent-Level Model Architecture Results:} We study which model architecture results in the best prediction accuracy across 6 models. For our specific robot model, which utilizes a single ToF sensor, we show that a Temporal convolutional neural network (TCN) performs the best.
    }
    \label{fig:train}
\end{figure}

\subsection{SBC Prediction Model}
 
In this section, we outline the data collection process, training, and design decisions for the SBC prediction model. The trained SBC model will be used by each robot in the swarm to estimate collective behavior from local perception. 

To build the training dataset, we use 8 differential-drive robots modeled and simulated in Isaac Sim, each outfitted with a ToF (time-of-flight) sensor that continuously measures line-of-sight distance over time. We run a suite of known swarm controllers, each associated with a known resulting emergent behavior (Cyclic Pursuit, Aggregation, Dispersal), obtained from prior robot swarms research~\cite{mattson2025discovery, mattson2023leveraging, brown2018discovery}. 

During simulation, local time-varying observation measurements are recorded from each robot once the swarm has converged to a steady behavior pattern. We address two prediction objectives: \textit{1) Behavior Convergence Detection:} Determining whether the swarm has converged to any stable behavior and \textit{2) Behavior Classification:} Determining the specific behavior to which the swarm has converged. \textcolor{red}{Once the ToF sensor streams are recorded, each sequence is labeled with the controller's corresponding one-hot encoded behavior tag and appended to the dataset for supervised training.} \textcolor{black}{Each controller rollout defines an episode-level label $\ell_e$. After behavior stabilization, each robot's ToF sequence is segmented into length-$T$ windows
$\mathbf{O}^{(e)}_{i,t}$, and each window inherits the corresponding
episode label, yielding samples $(\mathbf{O}^{(e)}_{i,t},\ell_e)$. The stabilized observation streams are then partitioned into disjoint 70/15/15\% training, validation, and test segments before window extraction, ensuring that no observation samples or windows are shared across splits.}


\begin{figure}
    \centering
    \includegraphics[width=\linewidth]{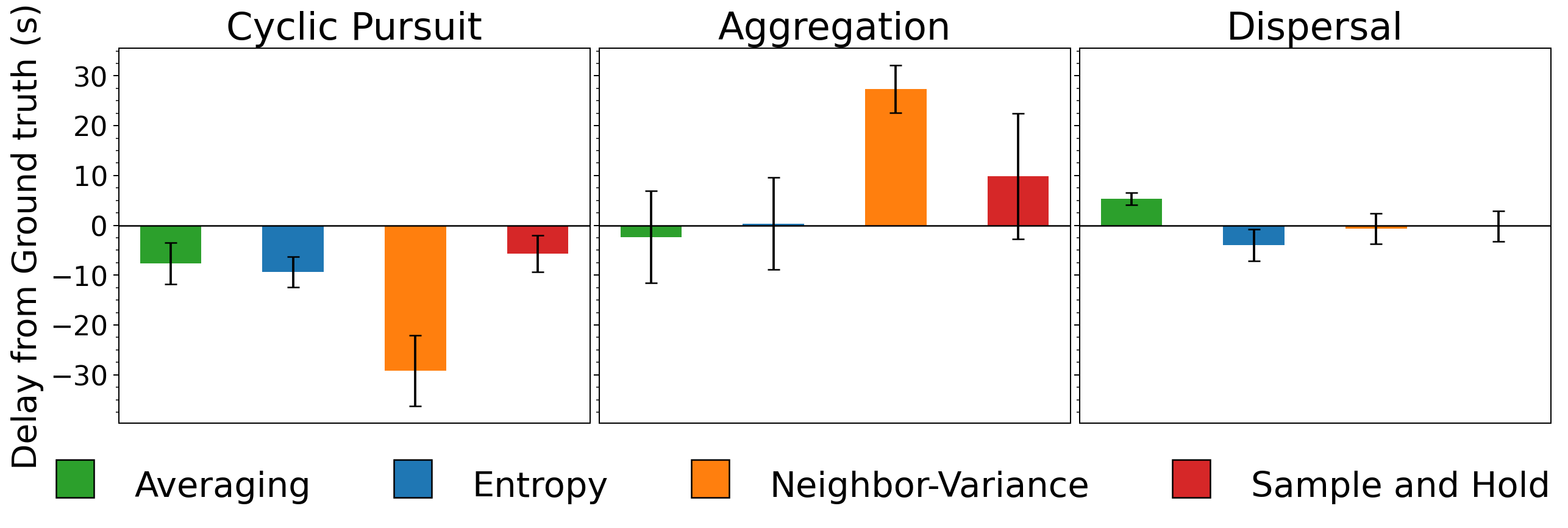}
    \caption{\textbf{Internal Belief Formation Results (Stage-1):} Across 3 swarm controllers, we measure the delay in full-swarm consensus across 4 methods compared to a ground truth measurement of behavior convergence (represented by 0). Negative results represent late predictions and Positive results represent early predictions. Results show the mean and standard error across 3 initial starting conditions.}
    \label{fig:bf}
\end{figure}
\begin{figure}
    \centering
    \includegraphics[width=\linewidth]{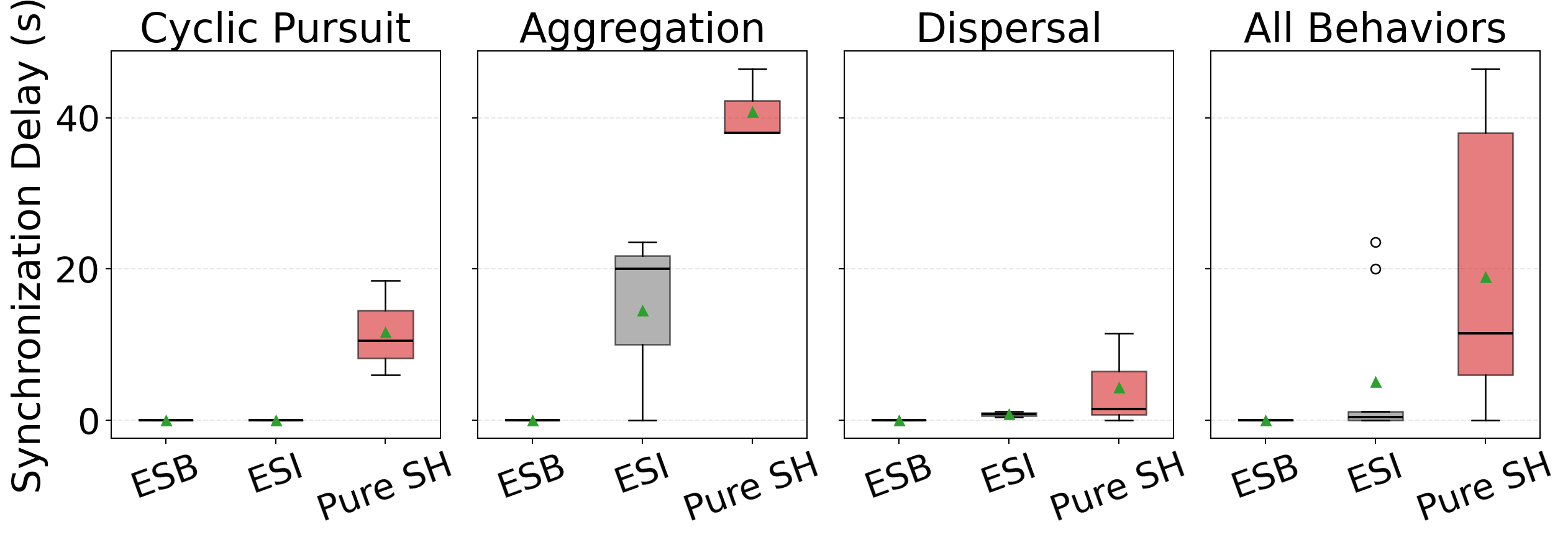}
    \caption{\textbf{Synchronization Error Results (Stage-2):} We compare 4 synchronization methods and measure the delay between the first robot forming its synchronized belief and the last robot. Across all methods, we find that combining Event-triggered, Sample-and-Hold with Boolean Gossip or IBS (ESI,ESB) leads to $<$5s delay across all behaviors. Results show the mean and standard error across 3 initial starting conditions.}
    \label{fig:slatency}
\end{figure}

During inference, the input is a sliding window of the ToF sensor stream. Longer window sizes provide a richer signal for behavior classification, but also delay the prediction process. Rather than reducing the window size, we retain a larger input size by increasing the sliding window stride.
\subsection{Neural Network Architecture Evaluation}\label{gt}

To select the best neural network architecture, we train a supervised learning model across multiple well-known neural network architectures using a cross-entropy loss. In our experiment, we evaluate GRU, 1D-CNN, TCN, RNN, LSTM, and Transformer models~\cite{modelsnn} using multiple metrics, including accuracy, macro F1 score, Brier, and expected calibration error (ECE)~\cite{brierece}. Figure~\ref {fig:train} shows the predictive quality and calibration error metrics for different architectures. We found that the Temporal Convolutional Network (TCN) performed best across all metrics.
\textcolor{red}{We attribute this to the ability of convolutional models to capture localized temporal patterns and extract meaningful feature representations from input signals, leading to improved calibration between behaviors. Additionally, TCN supports flexible input window sizes, which improves robustness to variations and noise in sensor signals, making it particularly well-suited for our setting.} 
\textcolor{black}{The TCN processes a $T\times d_o$ input using five residual causal
Conv1D blocks with 128 filters, kernel size 3, dilation factors
$\{1,2,4,8,16\}$, and dropout $0.2$, with $1\times1$ residual
projections when needed. The head uses global average pooling,
dropout $0.3$, Dense$(128,\mathrm{ReLU})$, dropout $0.3$, and an
$M$-class softmax output.}

\subsection{Stage-1 (Internal Belief Formation) Setup and Evaluation}
 
In this stage, we need to choose an appropriate Belief formulation mechanism that realizes its internal belief only after the swarm has converged to a stable collective behavior. Premature internal belief formation leads to inconsistent representations of true collective behavior.

As \textcolor{red}{a ground-truth metric}\textcolor{black}{a proxy ground-truth}, we use learned thresholds on swarm-level measures of \textit{scatter}, \textit{radial variance}, and \textit{average speed} from~\cite{brown2018discovery}. A human observer annotates the time of visual stabilization for a subset of controller runs, which are then used to calibrate the threshold parameters. These learned thresholds then automatically label the behavior convergence time for other controllers. 

To ensure a fair comparison across methods, we optimize the parameters of each consensus strategy using a hyperparameter search library, Optuna~\cite{ozaki2025optunahub}. 
We tune the parameters of each strategy to minimize the delay between the internal belief formulation and the ground-truth convergence time, while penalizing early convergence. 

The results in Figure~\ref {fig:bf} illustrate the temporal gap between the ground-truth detection of a swarm behavior and the point at which all agents set their internal belief to that same behavior. Our results show that the Sample-and-Hold method achieves the lowest combined delay compared to other consensus strategies. Averaging performs next best, showing competitive but slightly higher delay.

\subsection{Stage-2 (Synchronization) Setup and Evaluation}
We adopt Sample-and-Hold (SH) as the baseline belief-formation strategy for local internal belief updates and augment it with synchronization mechanisms. Specifically, we compare three methods: (1) Pure SH, (2) ETC combined with SH and Boolean Gossip synchronization (ESB), (3) ETC combined with SH and Integrated Belief Synchronizer (ESI). For each method, we evaluate synchronization delay and communication overhead, measured as the total message count (until convergence), to analyze the trade-off between synchronization and communication efficiency. 

\begin{figure}[t]
    \centering
    \includegraphics[width=\linewidth]{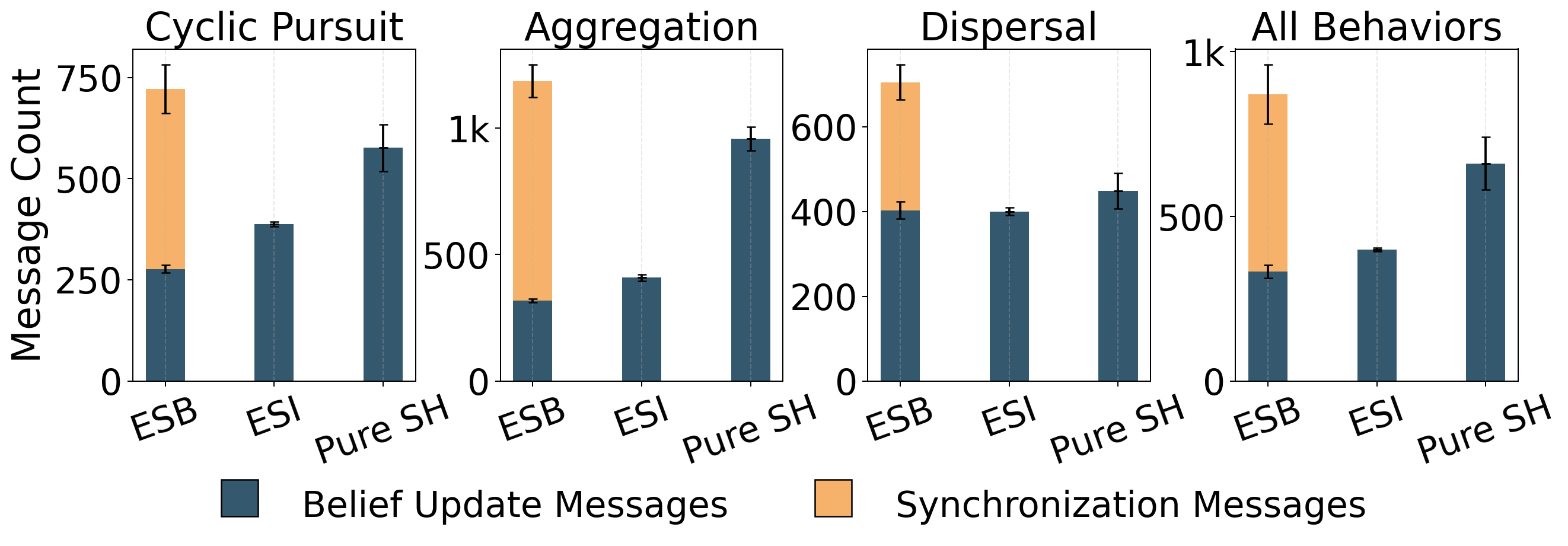}
    \caption{\textbf{Message Count Results:} For constrained deployments with a fixed communication budget, we examine the number of total messages transmitted under each of the different synchronization methods. Across all methods, we find that combining Sample-and-Hold with Integrated Belief Synchronizer (ESI) leads to the lowest number of required messages across all behaviors. Results show the mean and standard error across 3 initial starting conditions.}
    \label{fig:messagec}
\end{figure}

In Figures~\ref{fig:slatency} and~\ref{fig:messagec}, we observe that ESI performs slightly worse in the aggregation behavior compared to ESB and Pure SH. This is due to the formation of multiple groups, where the communication radius of one group does not overlap with that of another, resulting in higher synchronization delays. This shows that ESI can have lower communication overhead and maintain good synchronization, while ESB can achieve much better synchronization at a higher communication cost.

We then evaluate SyncSBC with varying swarm sizes and communication radii, including 5, 8, and 16 robots. We chose these sizes to evaluate whether our SBC prediction model trained on 8 robots can work for both smaller swarms and swarms with approximately double the number of agents. Our objective is to evaluate how communication and synchronized belief formation depends on swarm size and communication radius. 

\begin{figure}[t]
    \centering
    \includegraphics[width=0.9\linewidth]{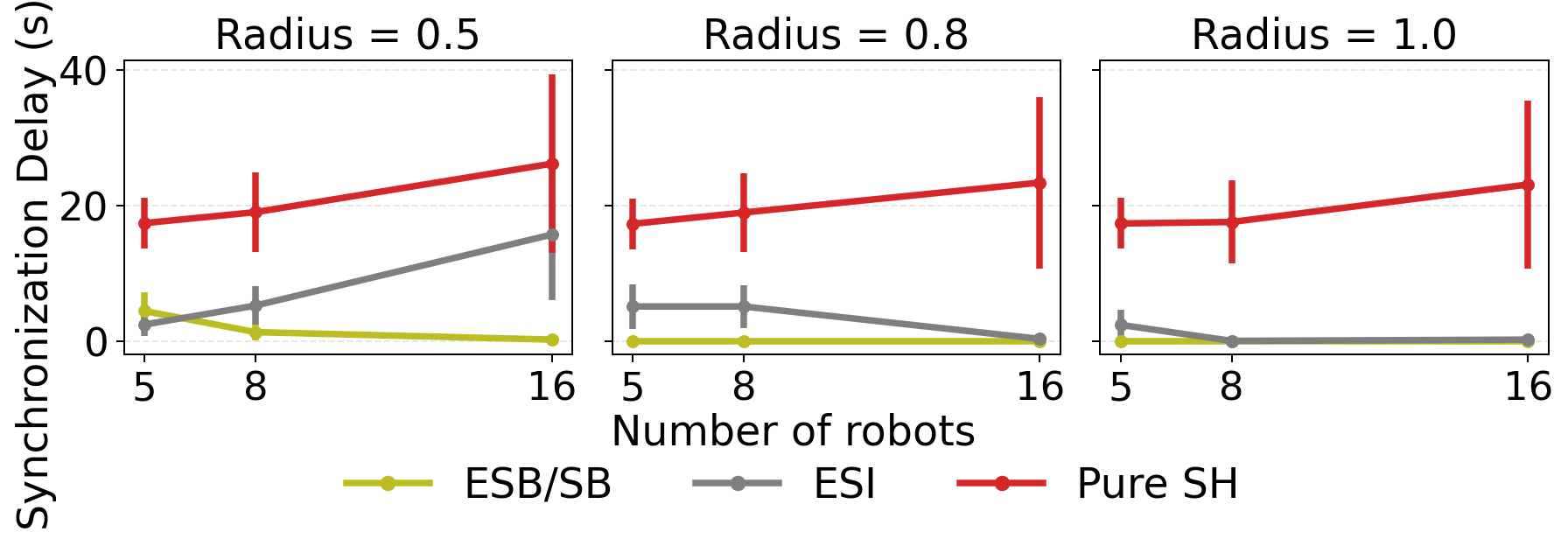}
    \includegraphics[width=0.9\linewidth]{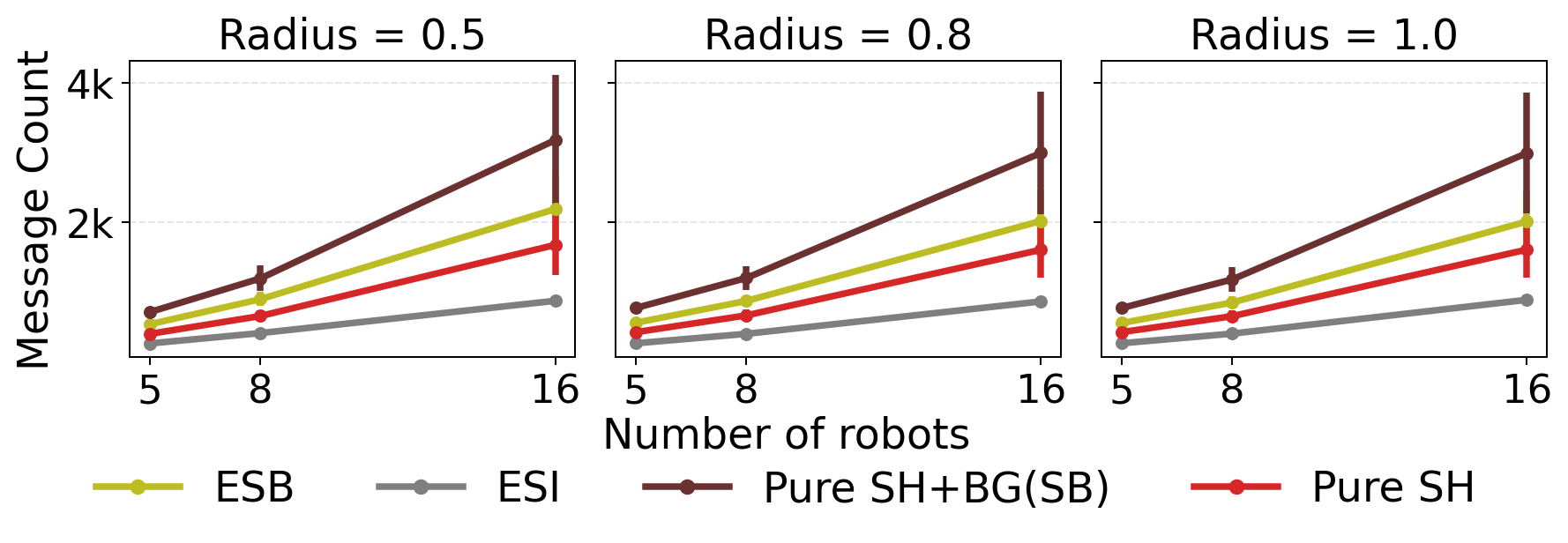}
    \caption{\textbf{Robots vs Radius:} We examined SyncSBC on swarms with varying number of robots and communication radii. We find that ESB and ESI achieve the lowest synchronization delay, especially in domains with many robots. While ESB has a higher communication rate, ESI performs well when connectivity within the swarm is dense, while minimizing communication. Results indicate mean and standard error over 9 runs.
    }
    \label{fig:radii}
\end{figure}

We test this by showing synchronization delay and message counts. We see in Figure~\ref{fig:radii} that increasing the communication radius $r_c$ improves synchronization. Synchronization delay is closely tied to the dynamic connectivity of the swarm; a larger neighbor set per robot allows faster message propagation, allowing robots to exchange beliefs with more peers before concluding on a behavior. Connectivity is also influenced by swarm size. Increasing the number of robots in a swarm increases network density, thereby further improving synchronization performance. This effect is evident in ESI, where even with a fixed radius of 0.8m, a larger swarm achieves better synchronization at a lower communication rate. In contrast, ESB requires more communication to achieve synchronization as swarm size increases. We can also observe that event-triggered communication maintains comparable synchronization while significantly reducing communication cost.


\begin{figure}
    \centering
    \includegraphics[width=\linewidth]{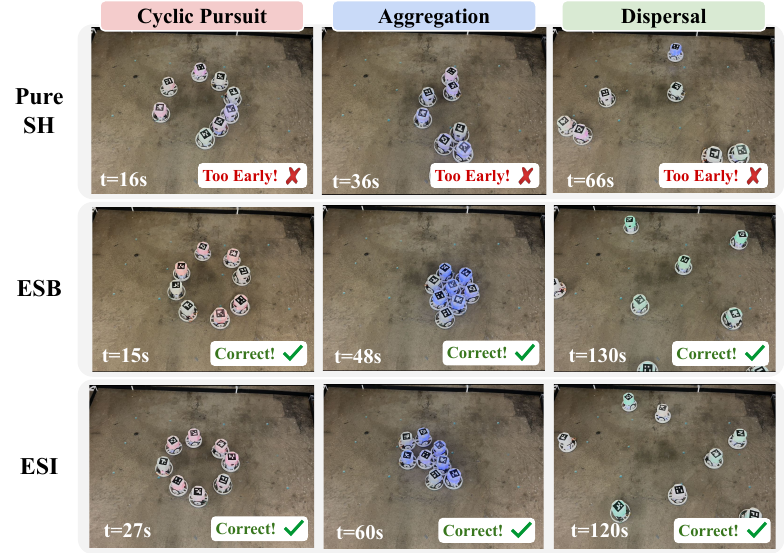}
    \footnotesize    
    \textbf{Synchronization Delay ($\Delta$)}
    \setlength{\tabcolsep}{4pt}
    \begin{tabular}{lcc}
    \toprule
     & \textbf{Cyclic Pursuit$\rightarrow$Aggregation} 
     & \textbf{Aggregation$\rightarrow$Dispersal} \\
    \midrule
    \textbf{Pure SH} & 18.0001 s & 26.0000 s \\
    \textbf{ESB}     & 0.0927 s  & 0.0927 s  \\
    \textbf{ESI}     & 1.9947 s  & 0.0000 s  \\
    \bottomrule
    \end{tabular}
    \normalsize
    
    \caption{\textbf{Application: Behavior Switching.} (top) Using only decentralized SyncSBC, swarm robots must synchronize a controller switch from Cyclic Pursuit (left) to Aggregation (middle) to Dispersal (right). (bottom) The synchronization delay for two controller switches, with each method having separate runs with the same initialization conditions. Pure SH often switches controllers too early (before consensus) and suffers from major synchronization delay. compared to ESB and ESI. }
    \label{fig:rbsw}
    
\end{figure}

\begin{figure}[t]
    \centering
    \includegraphics[width=0.85\linewidth]{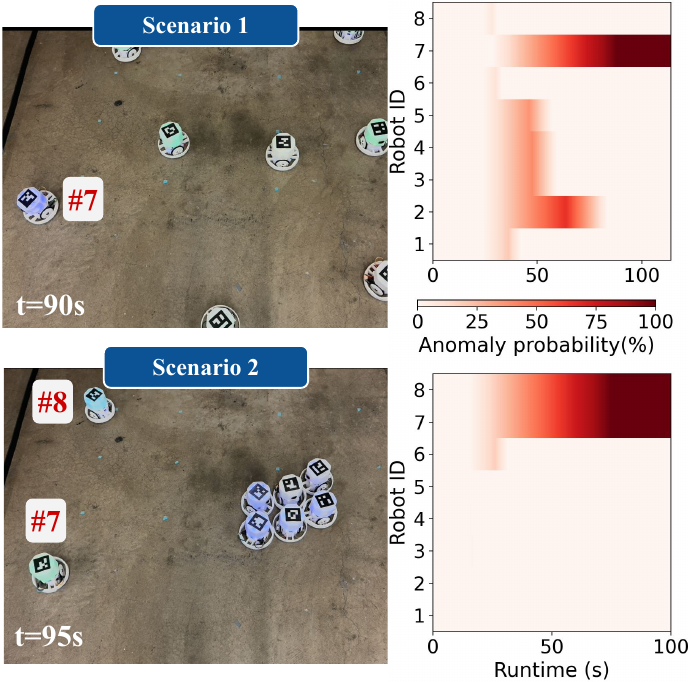}
    \setlength{\abovecaptionskip}{-1pt}
    \caption{\textbf{Application: Anomaly Detection.} (left) SyncSBC provides a mechanism where robots can self-inform themselves of incorrect behavior, i.e., anomaly detection. We studied two anomaly detection scenarios where 1-2 robots' behaviors have been deliberately halted or modified during swarm execution.  (right) The likelihood of anomaly as a function of runtime, where the robots (robot \#7 in scenario 1, \#7 and \#8 in scenario 2) increase their probability over time.}
    \label{fig:rad}
\end{figure}
\section{Closed-Loop Control on Real Robots}

To demonstrate the practical applicability of SyncSBC on real robots, we use differential drive HeRo+ robots~\cite{mattson2025discovery}, shown in Figure~\ref{fig:robot}(a), an extension of the open source HeRo~\cite{rezeck_hero_2023}.  
Here, we showcase two possible use cases: 

\subsection{Autonomous Behavior Switching}
By coupling synchronized consensus with decentralized control, coordinated swarm behaviors are possible without reliance on centralized supervision or a centralized controller. One such example is behavior switching. We define autonomous swarm behavior switching as a sequenced behavior-switching technique that switches between a global set of predefined behavior controllers, ensuring behaviors are in sync across the swarm. To implement a behavior switch, each robot uses its own synchronized belief signal $s_i$ from SyncSBC as a trigger to the next behavior controller. Here, we tested switching from Cyclic Pursuit$\xrightarrow{}$Aggregation$\xrightarrow{}$ Dispersal; the results are visualized in Figure~\ref {fig:rbsw}. We deploy, (1) Pure SH, (2) ESB, and (3) ESI on separate runs. 

The results show that naively forming a global belief via basic consensus methods is insufficient for tasks that require tight synchronization under dynamic conditions. Pure SH (Sample-and-Hold) performs poorly at maintaining consistent, synchronized transitions. Although robots switch with lower delay in the first transition, synchronization errors compound with subsequent transitions, resulting in higher variance and misalignment within the swarm.
In contrast, ESB (Event-triggered, Sample-and-Hold, Boolean gossip) and ESI (Event-triggered, Sample-and-Hold, IBS) exhibit clearly recognizable, stable behaviors before transitioning to the next behavior, while minimizing synchronization delay. While ESI shows a slightly higher synchronization delay for the first transition, the swarm quickly aligns itself for the subsequent transition. These results highlight synchronization as a critical enhancement to consensus-based strategies, mainly in groups where neighborhood alignment is essential. 
 
\subsection{Anomaly detection}
SyncSBC's pipeline incorporates a distinctive mechanism within its synchronization stage. In this phase, each robot either propagates a ``not ready'' signal through Boolean gossip (ESB) or compares with the largest member set in the Integrated Behavior Synchronizer (ESI). 

Anomaly likelihood increases when robot $i$ strongly rejects behavior $k$ ($b_{i,k}=0$ and $\argmax_k P_{i,k}\neq k$) while others propagate $s_{i,k}=1$ in ESB. Similarly, under ESI, the anomaly likelihood increases when the robot's internal belief $b_i$ repeatedly disagrees with the largest membership set $\mathcal{M}_k$. This allows individual agents in the swarm to identify their inconsistencies or misalignment during synchronization.  

To demonstrate anomaly detection, we run two scenarios. In Scenario 1, a single robot is halted while the rest of the swarm executes a dispersing behavior. In Scenario 2, we set two robots to use modified behavior controllers different from the majority. The results in Figure~\ref{fig:rad} show probability estimates for each individual robot's self-assessment of their anomaly likelihood. In Scenario 1, Robot \#7 is flagged by the system, with its anomaly likelihood increasing over time as it remains misaligned with the dispersing swarm. Similarly, in Scenario 2, Robots \#7 and \#8 exhibit consistently high anomaly likelihood, reflecting their persistent deviation from the majority.
These experiments demonstrate that SyncSBC not only enables synchronized consensus but also provides an inherent mechanism for real-time anomaly detection, providing the potential for members of a swarm to identify, isolate misaligned agents, and self-heal without external supervision and in a fully decentralized manner.

\section{Conclusion}
We introduced SyncSBC, a novel framework consisting of real-time behavior classification using \textit{only the local observation history from individual members of the swarm} and a decentralized consensus mechanism that enables synchronized swarm action. We show that our SBC model achieves $>$95\% accuracy on held-out testing data, and swarm-level consensus with $<$1s average synchronization delay in simulation, compared to $>$10s delay seen in agreement-only consensus methods. On a real robot swarm consisting of 8 HeRo+ robots, our method achieves $<$3s delay versus $>$15s with agreement-only strategies. Our work also demonstrates that SyncSBC can successfully be applied to two important swarm problems---Behavior Switching and Anomaly Detection. 
Future work should consider the impact of sensing modality, state observability, and behavior complexity on the SyncSBC pipeline as well as applications to self-healing behaviors and environmental decision making. Overall, our work highlights promising results for fully decentralized classification, consensus, and synchronization for communication-constrained real-world swarm robotics.



\bibliographystyle{plain}
\bibliography{references}

\end{document}